\documentclass{article}

\usepackage{graphicx}
\usepackage{float}
\usepackage{enumitem}    
\usepackage{subcaption}
\usepackage{booktabs}
\usepackage[bottom]{footmisc}
\usepackage{hhline}
\usepackage{comment}
\usepackage{fancyhdr}

\newcommand{\squeezeup}{\vspace{-2.5mm}}

\usepackage{hyperref}

\usepackage[accepted]{icml2018}

\icmltitlerunning{Modeling Mistrust}

\begin{document}

\twocolumn[
\icmltitle{Modeling Mistrust in End-of-Life Care}




\begin{icmlauthorlist}
\icmlauthor{Willie Boag}{mit}
\icmlauthor{Harini Suresh}{mit}
\icmlauthor{Leo Anthony Celi}{mit}
\icmlauthor{Peter Szolovits}{mit}
\icmlauthor{Marzyeh Ghassemi}{mit,to,vec,ver}
\end{icmlauthorlist}

\icmlaffiliation{to}{University of Toronto}
\icmlaffiliation{vec}{Vector Institute}
\icmlaffiliation{ver}{Verily}
\icmlaffiliation{mit}{MIT}

\icmlcorrespondingauthor{Willie Boag}{wboag@mit.edu}

\vskip 0.3in
]

\printAffiliationsAndNotice{}




\begin{abstract}
In this work, we characterize the doctor-patient relationship using a machine learning-derived trust score. We show that this score has statistically significant racial associations, and that by modeling trust directly we find stronger disparities in care than by stratifying on race. We further demonstrate that mistrust is indicative of worse outcomes, but is only weakly associated with physiologically-created severity scores. Finally, we describe sentiment analysis experiments indicating patients with higher levels of mistrust have worse experiences and interactions with their caregivers. This work is a step towards measuring fairer machine learning in the healthcare domain. 
\end{abstract}

\section{Introduction}

There are well-established gaps in the American healthcare system for minority populations.
Groups that have been historically marginalized have also had worse treatment options and longitudinal health outcomes. 
Biases are especially troubling in the context of machine learning applied to clinical data. Bias can be replicated and exacerbated in the model's future recommendations \citep{ensign2017runaway}. For example, black and Hispanic patients are often given less pain medication for equivalent injuries and reported pain levels \citep{jama:pain-meds-kids,plos:pain-meds}. If this pattern is present in the training data for a model built to recommend treatment, it would learn to associate race with pain medication dosage.

Differences in care have also been established during end-of-life (EOL), when critically ill patients are confronting death \citep{muni:race-ses-eol-icu,lee:dying-icu}. Previous work has suggested that medical disparities can reflect higher levels of mistrust for the healthcare system among black patients. It is said that blacks are more suspicious of the clinical motives in advance directives and do-not-resuscitate (DNR) orders \citep{wunsch2010three}, and believe that the healthcare system was controlling which treatments they can receive \citep{perkins2002cross}. When the doctor-patient relationship lacks trust, patients may believe that limiting any intensive treatment is unjustly motivated, and demand higher levels of aggressive care. While there are clinical examples of exemplary end-of-life care, studies have highlighted that aggressive care can lead to painful final moments, and may not improve patient outcomes \citep{cipolletta2014good}.


Prior works in the FATML community have attempted to mask out features that may lead to disparate treatment \citep{zemel2013learning}, but including information about race may be important for some clinical tasks (e.g., if there are differences in recommended care by genetic makeup). In such a setting, quantifying bias and establishing proxy measures for medical trust is particularly important. 

In this work, we present three contributions. 
\begin{itemize}[noitemsep,topsep=0pt,parsep=0pt,partopsep=0pt]
\item We present a trust metric derived from coded doctor-patient interactions. 
\item We demonstrate that our trust score captures treatment differences by showing disparities in end-of-life care are pronounced when patients are stratified on trust.
\item We validate our mistrust metric using sentiment analysis of patients' notes.
\end{itemize}


For further analysis -- including the analysis of three different proxy scores for trust -- see \cite{boag-eol-mistrust}.

\section{Background and Related Work}
The quantity of health-related data is increasing rapidly, from genetic data to medical images like x-rays \citep{kruse2016challenges,raghupathi2014big}. These rapid advancements have facilitated large-scale machine learning methods to guide care. \citeauthor{ferryman2018fairness} give an overview of fairness issues that may arise with such advances in personalized medicine. However, further research into these risks and the feasibility of applying existing FATML work to healthcare domains is limited.



Socialized mistrust of the medical community in minority groups has been established as a factor in care differences \citep{medicalapartheid}. Family members of African American patients are more likely to cite absent or problematic communication with physicians about EOL care \citep{hauser:advance-directives}. Similarly, in surveys, African Americans report lower rates of satisfaction with the quality of care that they received by physicians \citep{hanchate:minorities-cost-more}. In end-of-life care, a mistrustful patient could be more resistant to a doctor's recommendation of comfort-based care, and instead insist receiving all possible treatments even if they are overly aggressive \citep{garrett:lifesustaining,hopp:race-eol}.

Trust is difficult to quantify, and shaped by subtle interactions such as perceived discrimination, racial discordance, poor communication, language barriers, unsatisfied expectations, cultural stigmas and reputations, and more \citep{cultural-mistrust}. Trust is very important to success of a hospital stay; previous work has found that increased levels of doctor-patient trust were associated with stronger adherence to a physician's advice, increased patient satisfaction and improved health status \citep{trust-improves-outcomes}.

Previous efforts to create trust-based measures that correlate with outcomes have relied on surveys, which can be difficult to conduct for both theoretical (selection bias) and practical (cannot be done for retrospective, de-identified data) concerns \citep{lee:dying-icu}.

\section{Data}
\label{sec:data}

We use the publicly-available Medical Information Mart for Intensive Care (MIMIC-III) v1.4~\citep{johnson2016mimiciii}. This database contains de-identified EHR data from over 58,000 hospital admissions for nearly 38,600 adult patients. The data was collected from Beth Israel Deaconess Medical Center from 2001--2012. We examine a cohort of black and white patients in end-of-life care.
We examine patients who have a hospital stay which lasted at least 6 hours, and have either died in the hospital, were discharged to hospice, or were discharged to a skilled nursing facility\footnote{This was done because the notes indicate some SNF patients are discharged on hospice without coding that into the EHR.} (SNF). These experiments are repeated on a stricter definition of an EOL cohort (which excludes SNF patients) in \cite{boag-eol-mistrust}, shows the same trends as this work, but with less statistical power because of smaller sample sizes.
Both our data extraction and modelling code are made available\footnote{\url{https://github.com/wboag/eol-mistrust}} to enable reproducibility and further study \citep{johnson:mlhc17-reproduce}.

In order to measure disparities in aggressive end-of-life procedures, we extracted treatment durations (in minutes) from MIMIC's derived mechanical ventilation (\textit{ventdurations}) and vasopressor (\textit{vasopressordurations}) tables \footnote{Available freely at https://github.com/MIT-LCP/mimic-code/tree/master/concepts/durations.}. Due to the noisiness of clinical measurements,\footnote{for instance, when one treatment span is erroneously coded as two back-to-back smaller spans} we merge any treatment spans that occurred within 10 hours of each other.\footnote{This heuristic was suggested by MIMIC staff because 10 hours is approximately the shift of a nurse, and treatment duration events might get recorded once at the beginning of each shift.} If a patient had multiple spans, such as an intubation-extubation-reintubation, then we consider the patient's treatment duration to be the sum of the individual spans.

In this work, we wish to better understand and quantify the nuances of a patient's interactions with their nurses and doctors. We accomplish this using two sources: clinical notes and coded chart events. We obtain the notes of every patient who had a stay of at least 12 hours in the ICU. This resulted in 48,273 admissions and over 800,000 notes. Most notes are nursing notes, discharge summaries, physician notes, and social worker notes. To supplement this narrative prose, we also extract coded information from the MIMIC \textit{chartevents} table, which records many interpersonal aspects of the patient's stay, including: code status, health literacy (e.g. whether there is a healthcare proxy), behavioral and mental status assessments, family communications, pain management, whether the patient was restrained, whether the patient wanted help bathing, support services, and more.

\section{Methods and Experiments}


We aim to replicate previously demonstrated racial disparities in end-of-life care using MIMIC-III~\citep{johnson2016mimiciii}. We take as reference a set of three recent papers which examined the racial disparities in end-of-life care for nonwhite or minority populations \citep{ices:immigration,muni:race-ses-eol-icu,lee:dying-icu}. We compared the differences of patient outcomes between white and black populations using Mann-Whitney analysis for non-normally distributed variables (treatment durations, mistrust metric scores). In accordance with prior work, we consider p-values $<.05$ to be statistically significant.

\subsection{Establishing a Medical Mistrust Metric}\label{mistrust-methods}
 
Ideally, the gold standard for measuring trust would be a survey where the patient -- in their own words -- describes their feelings and relationship with their caregiver \citep{trust-improves-outcomes}. However, such surveys were not recorded for MIMIC patients. But we believe that there is still a useful signal of the trust which can be inferred from the EHR. We quantify mistrust in a novel way by seeding a supervised machine learning task with labels which serve as a proxy for mistrust.
Our goal is to model the underlying relationship and create a trust score which aims to explain treatment disparities better than race does.

We extract coded interpersonal features from the \textit{chartevents} table for all MIMIC patients.
Some of the information extracted includes: indication of family meetings, patient education, whether the patient needed to be restrained, how thoroughly pain is being monitored and treated, healthcare literacy (e.g. whether the patient has a healthcare proxy), whether the patient has a support system (such as family, social workers, and religion), and agitation scales (Riker-SAS and Richmond-RAS). In total, we extract 620 unique binary indicators.

We use a simple rule-based search through the notes to determine
whether the patient has non-compliance documented somewhere in their notes (e.g. medical advice, regimens, follow-ups, etc.).
Noncompliance indicates a very overt mistrust; rather than just holding an unspoken resentment, the patient actually defies their doctor's orders. Because crossing this line explicitly demonstrates that the patient is willing to disregard physician decisions, it is a suitable prediction target for training the model to quantify mistrust. 

Of the 48,273 hospital admissions, we find 464 with notes that document noncompliance, and fit an L1-regularized Logistic Regression\footnote{\url{http://scikit-learn.org/stable/modules/generated/sklearn.linear_model.LogisticRegression.html}} model with chartvents features to predict whether the patient was noncompliant. Once the model is trained, we use the classifier's predicted probability for a new patient as a proxy for their degree of mistrust.

\subsection{Validating the Mistrust Metric}

Throughout a patient's stay, caregivers write narrative prose notes to document administered care, family meetings, patient preferences, reminders, warnings, and the patient's quality of care. 
In documenting their impressions of how to best understand and interact with their patients, caregivers can give clues into their relationship with the patient and family. 

Clinical notes have been used for prediction tasks in previous work \citep{ghassemi2014unfolding} but not for investigating mistrust. Sentiment analysis of clinical notes has also been used to measure whether one group of patients has a better experience, on average, than another group \cite{roy:sentiment}. We use the Pattern software package for sentiment analysis \cite{pattern}. 

For a given hospital admission, we compute the sentiment score of the concatenation of that patient stay's notes. Once we compute the score for each stay in our population, we scale the distribution of scores to be zero-mean and unit-variance in order to give better sense to the differences in sentiment, relative to average. Of particular interest is the differences in sentiment between different groups: white-and-black; trustful-and-mistrustful. As a sanity check, we hypothesize that sick patients have more negative stays than healthy patients, so we also stratify into low- and high-risk subpopulations using the Oxford Acute Severity of Illness Score (OASIS) score \cite{johnson:oasis}. We create subpopulations to be same size as black/white split, i.e. the white:black dataset size ratio is the same as the trustful:mistrustful dataset size ratio.

\section{Results}
\subsection{Creation of a Mistrust Metric}

\begin{table}
  \begin{center}
    \caption{Top-3 most positively and negatively informative chartevent features for tuning the mistrust metric.}
    \label{tab:metric-weights}
    \hspace*{-1cm}
    \begin{tabular}{|c|c|}
    	\hline
                     \textbf{feature} & \textbf{weight} \\ \hline \hline
                         state: alert & -1.0156  \\ \hline
            riker-sas scale: agitated &  0.7013  \\ \hline
                           pain: none & -0.5427  \\ \hline
richmond-ras scale: 0 alert and calm  & -0.3598  \\ \hline
              education readiness: no &  0.2540  \\ \hline
         pain level: 7-mod to severe  &  0.2168  \\ \hline
    \end{tabular}
  \end{center}
\end{table}

Table~\ref{tab:metric-weights} shows the three most positively and most negatively informative weights used to predict a mistrust metric (Section \ref{mistrust-methods}). The features align well with our intuitive notion of mistrust: patients who are agitated and not receptive to education are more likely to be mistrustful, whereas calm, pain-free patients are more willing to trust their doctor.

We observe a statistically significant racial disparity in the mistrust metric, where the median black patient has a higher level of mistrust than the median white patient using the Mann-Whitney test (p=0.003). This is not surprising, given the extensive literature investigating differences in iatrophobia by race \citep{medicalapartheid}.

\subsection{Significant Disparities in EOL Care}
\label{sec:treatment-stats}

\subsubsection{Race-Based Disparities}

We demonstrate racial treatment disparities in the MIMIC dataset. Figure \ref{fig:treatments-race} highlights the differences in white and black populations for aggressive treatment durations. Figures \ref{fig:treatments-race}a and \ref{fig:treatments-race}b show that for both mechanical ventilation and vasopressors, the median black patient receives a longer duration of treatment, perhaps suggesting a reluctance to transition to palliative care. While these results only show statistical significance for ventilation, the same trends are also observable for vasopressor administration. 

\squeezeup

\subsubsection{Trust-Based Disparities}

\begin{figure}
\centering
    \caption{We observe racial disparities in for black patients (when compared to white patients) for the duration of aggressive interventions (vasopressors and ventilation). Medians are indicated by dotted lines; differences are significant ($p < 0.05$) for ventilation but not for vasopressors.}
    \begin{subfigure}{.45\linewidth}
      \centering
      \includegraphics[width=\textwidth]{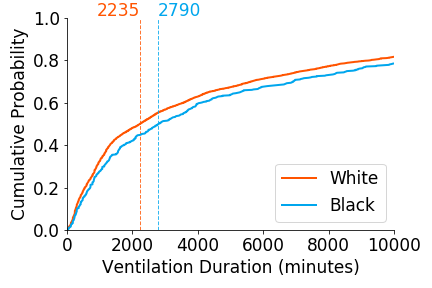}
      \caption{CDF of ventilation duration by race ($p=.005$). 
      }
    \end{subfigure}
	~
    \begin{subfigure}{.45\linewidth}
 		\centering
	 	\includegraphics[width=\linewidth]{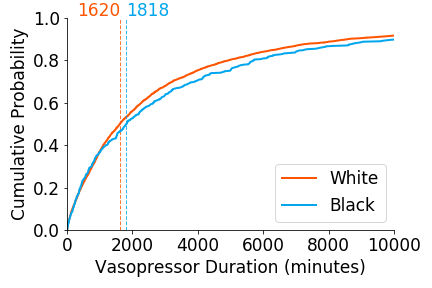}
 		\caption{CDF of vasopressor duration by race ($p = 0.12$).
      }
    \end{subfigure}
    \label{fig:treatments-race}
\end{figure}

\begin{figure}
  \caption{Stratifying patients by mistrust directly shows starker disparities in care than race. Medians for ventilation and vasopressor durations are indicated with dotted lines and all differences between the groups are significant. }
\centering
    \begin{subfigure}{.45\linewidth}
      \centering
      \includegraphics[width=\textwidth]{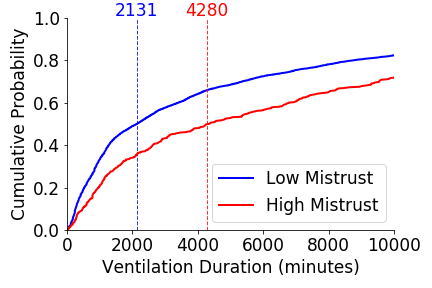}
      \caption{CDF of ventilation duration ($p<.0001$).\\
      }
    \end{subfigure}
	~
    \begin{subfigure}{.45\linewidth}
 		\centering
	 	\includegraphics[width=\linewidth]{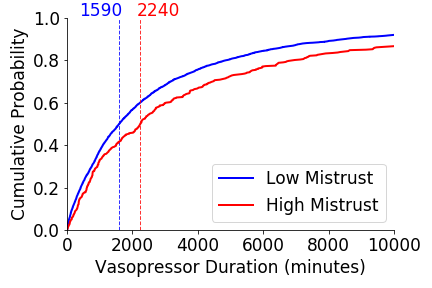}
 		\caption{CDF of vasopressor duration ($p<.0001$). \\
      }
    \end{subfigure}    
    \label{fig:treatments-trust}
\end{figure}

Using the mistrust metric, we can rank the patients by trust score and stratify them into two groups: low- and high-mistrust. Figure \ref{fig:treatments-trust} revisits the experiments from Figure \ref{fig:treatments-race} except stratified into low and high mistrust instead of white and black populations.\footnote{For each treatment, we preserve the same size difference of stratified groups in order to maintain consistency in sample sizes for significance testing, e.g,. because the black group contains 510 patients for ventilation, we compare the 510 lowest trust patients against the 4811 highest trust patients.} We can see from Figure \ref{fig:treatments-trust}b that trust-based disparities in vasopressor durations are significant. The difference between medians of each group is 650 minutes for vasopressors (whereas the difference stratified by race was 200 minutes). This gap is even larger for ventilation durations, as shown in Figure \ref{fig:treatments-trust}a: the trust-based stratification shows a 2150-minute difference between medians, in contrast to the 550-minute gap for the race split in Figure \ref{fig:treatments-race}a.

\subsection{Low Trust Patients Have The Most Negative Notes}

Table \ref{tab:sa-results} shows the differences in sentiment analysis scores between race, severity of illness, and trust.\footnote{Note that a naive application of tokenization is misleadings, as even positive notes containing "Date:[**5-1-18**]" were tagged as negative because the tool's string-matching algorithm was identifying ``:['' as negative emoticon use.} As a reminder, the scores were normalized to be zero-mean and unit-variance. It is interesting that every subpopulation (and indeed the full population) median score is at least slightly negative, indicating that the distribution has a positive skew.

We observed statistically significant differences in the population means ($p<.05$) for all three stratifications using the Mann-Whitney test. In particular, we see that black patients, high risk patients, and low trust patients all have stronger levels of negative sentiment in their notes. However, the low trust cohort had the most extreme negative sentiments. The median low trust sentiment (-0.242) was more than twice as far from the center as the median black sentiment (-0.110), further suggesting that the mistrust metric is able to tease out the cases with poor caregiver interactions and impressions.

\begin{table}
  \begin{center}
    \center
    \caption{Median sentiment analysis of cohorts stratified by race, severity, and trust.\\}
    \label{tab:sa-results}
    \begin{tabular}{|c|c|c|}
    	\hline
                     \textbf{population} & \textbf{N} & \textbf{median} \\ \hline \hline
                         White         & 9629  & -0.064 \\ 
                         Black         & 1164  & -0.110 \\ \hline \hline
                         Low Severity  & 9629  & -0.058 \\ 
                         High Severity & 1164  & -0.167 \\ \hline \hline
                         High Trust    & 9629  & -0.049 \\ 
                         \textbf{Low Trust}     & \textbf{1164}  & \textbf{-0.242} \\ \hline \hline
    \end{tabular}
  \end{center}
\end{table}

\subsection{Not Just Some Severity Score Proxy}

One initial concern we had was that this mistrust metric might have actually been more similar to a severity score like OASIS than intended. Certainly, high-risk patients are treated differently than the general population. To dispel this concern, we compared the pairwise correlations between the mistrust score, OASIS, and SAPS II -- another severity measure \cite{sapsii}. Table \ref{tab:correlations} shows that the two well-established acuity scores, OASIS and SAPS II, have a strong correlation of 0.68. On the other hand, the mistrust score does not seem to simply recapitulating severity of illness, as indicated by its weak (0.095 and 0.045) correlations with the other two scores.

\begin{table}
  \begin{center}
    \caption{Pairwise Pearson correlations between severity scores and mistrust score.\\}
    \label{tab:correlations}
    \hspace*{-1cm}
    \begin{tabular}{|c||c|c|c|}
    	\hline
                 & OASIS & SAPS II & Mistrust \\ \hline \hline
        OASIS    & 1.0   & 0.680   & 0.095    \\ \hline
        SAPS II  & 0.680 & 1.0     & 0.045    \\ \hline
        Mistrust & 0.095 & 0.045   & 1.0      \\ \hline
    \end{tabular}
  \end{center}
\end{table}

\section{Limitations}
The primary limitation of this study is that the labels for tuning the weights for the mistrust metric were generated with a simple rule-based search for the word "noncompliant" in a patient's clinical notes. Not only does this narrow definition of mistrust fail to capture some of the more subtle interactions in unhealthy doctor-patient relationships, but it also could falsely attribute malice to logistic issues such as being noncompliant with home medications because of a lack of access to prescriptions. In practice, however, we did not observe many false positive examples, and the mistrust metric -- both in feature weights and in analysis of treatment/sentiment disparities -- indicates that it is a sufficient first-attempt proxy to capture the more difficult-to-measure quantity of "trust."

\section{Conclusion}
In this work, we demonstrate that black patients receive -- sometimes significantly -- longer durations of invasive treatments in the MIMIC database. Though these trends have been studied in private datasets, we present our replicable analysis on a public dataset.

We create a mistrust score by using coded interpersonal features to predict patient noncompliance. 
This mistrust metric is a better identifier than race to show
disparities in both end-of-life care and sentiment.
However, this score also indicates a higher level of mistrust held by black patients than white patients.

Medical machine learning is moving forward at an exciting pace; we hope that this work will be a step towards creating models of human physiology that serve everyone, and do not propagate existing disparities in care. In order to achieve that goal, we must make better efforts to measure and understand these disparities.

\bibliography{example_paper}
\bibliographystyle{icml2018}

\end{document}